\pdfoutput=1
\documentclass[11pt]{article}

\usepackage{acl}

\usepackage{algorithm}
\usepackage{algpseudocode}

\usepackage{color, colortbl}

\definecolor{LightCyan}{rgb}{0.88,1,1}

\usepackage{booktabs}
\usepackage{esrelation}
\usepackage{tikz}
\usepackage{graphics}

\usepackage{siunitx}
\usepackage{graphicx}
\usepackage{xcolor}
\usepackage{amsfonts}

\usepackage{times}
\usepackage{latexsym}
\usepackage{cleveref}
\crefname{figure}{Fig}{Figs}%
\crefname{algorithm}{Algorithm}{Algo}%

\newcommand{\vclip}{\textsc{SDS-CLIP}}
\usepackage[T1]{fontenc}
\usepackage[utf8]{inputenc}

\usepackage{microtype}

\title{Distilling Knowledge from Text-to-Image Generative Models \\ Improves Visio-Linguistic Reasoning in CLIP}

\author{
  \textbf{Samyadeep Basu\textsuperscript{1, *}},
 \textbf{Shell Xu Hu\textsuperscript{2}},
  \textbf{Maziar Sanjabi\textsuperscript{3}},
  \textbf{Daniela Massiceti\textsuperscript{4}},\\
  \textbf{Soheil Feizi\textsuperscript{1}}
  \\
 \textsuperscript{1}University of Maryland, College Park,
 \textsuperscript{2}Samsung AI,
\textsuperscript{3}Meta AI,
\textsuperscript{4} Microsoft Research
\\
\small{
  \textbf{Correspondence:} {sbasu12@umd.edu}
 }
  } 
\begin{document}
\maketitle
\begin{abstract}
Image-text contrastive models like CLIP have wide applications in zero-shot classification, image-text retrieval, and transfer learning. However, they often struggle on compositional visio-linguistic tasks (e.g., attribute-binding or object-relationships) where their performance is no better than random chance. To address this, we introduce SDS-CLIP, a lightweight and sample-efficient distillation method to enhance CLIP's compositional visio-linguistic reasoning. Our approach fine-tunes CLIP using a distillation objective borrowed from large text-to-image generative models like Stable-Diffusion, which are known for their strong visio-linguistic reasoning abilities. On the challenging Winoground benchmark, SDS-CLIP improves the visio-linguistic performance of various CLIP models by up to 7$\%$, while on the ARO dataset, it boosts performance by up to 3$\%$. This work underscores the potential of well-designed distillation objectives from generative models to enhance contrastive image-text models with improved visio-linguistic reasoning capabilities.
% Image-text contrastive models such as CLIP are useful for a variety of downstream applications including zero-shot classification, image-text retrieval and transfer learning. However, these contrastively trained vision-language models often fail on compositional visio-linguistic tasks such as Winoground with performance equivalent to random chance. In our paper, we address this issue and propose a sample-efficient light-weight distillation method called SDS-CLIP to improve the compositional visio-linguistic reasoning capabilities of CLIP. The core idea of our method is to use differentiable image parameterizations to fine-tune CLIP with a distillation objective from large text-to-image generative models such as Stable-Diffusion which are relatively good at visio-linguistic reasoning tasks. On the challenging Winoground compositional reasoning benchmark, our method improves the absolute visio-linguistic performance of different CLIP models by up to $7\%$, while on the ARO dataset, our method improves the visio-linguistic performance by upto 3$\%$.  
% Our method reinforces that carefully designed distillation objectives from generative models can be leveraged to extend contrastive image-text models with improved visio-linguistic reasoning capabilities. 
\end{abstract}

\section{Introduction}
In recent years, multimodal models like CLIP~\cite{radford2021learning} have excelled in tasks such as zero-shot classification, image-text retrieval, and image-captioning~\cite{slip, yu2022coca, blip, mokady2021clipcap}. These models are also crucial components in various state-of-the-art pipelines for tasks like segmentation and object detection~\cite{clip_seg_1, clip_seg_2, minderer2022simple, region_clip}. However, they struggle with visio-linguistic reasoning tasks, such as determining the spatial relationships between objects in an image~\cite{yuksekgonul2023when, huang2023structureclip}. Notably, CLIP's performance on the challenging Winoground~\cite{thrush2022winoground, diwan2022winoground}, a benchmark designed to assess visio-linguistic reasoning, is close to random chance. This shortcoming is attributed to CLIP's contrastive objective which prioritizes shortcuts for retrieval, and thus impacts its ability to understand fine-grained object details and their positions~\cite{diwan2022winoground, thrush2022winoground}.

\begin{figure}
    \hskip -0.4cm
  \includegraphics[width=8.4cm, height=6.0cm]{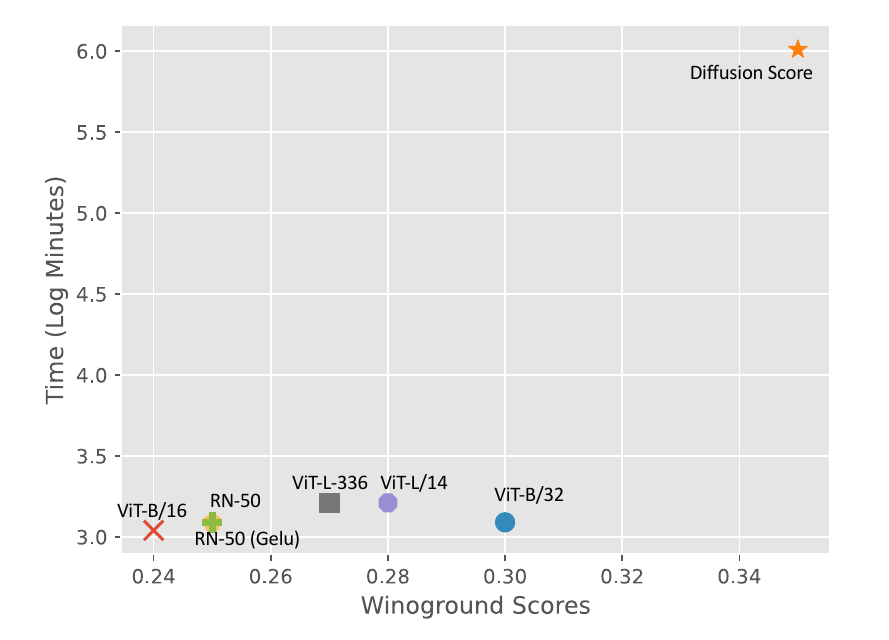}
    \caption{\label{diffusion_score} \textbf{CLIP variants underperform on Winoground, a visio-linguistic reasoning benchmark, compared to Diffusion Score from Stable Diffusion.} The diffusion score is computed from Stable Diffusion's loss function. Note that Diffusion Score takes $~18 \times$ more time than CLIP variants for inference (using 50 samplings during diffusion score computation). }%
    \vspace{-0.6cm}
\end{figure}

% In contrast, text-to-image generative models like Stable Diffusion~\cite{stable_diffusion} have been shown to exhibit strong visio-linguistic reasoning skills~\cite{li2023diffusion, clark2023texttoimage}, likely due to their text conditioning mechanism which enhances semantic consistency in their cross-attention maps.
% This has most recently been demonstrated by~\citet{li2023diffusion} on the Winoground benchmark. Here, they showed that the density estimates, or denoising diffusion scores, from Stable Diffusion can reliably be used to match captions to images, even when there are fine-grained differences in the spatial arrangements of objects. 
% This approach has similarly been demonstrated for other text-to-image generative models, including Imagen~\cite{clark2023texttoimage}, most of which outperform CLIP variants on Winoground's image-matching task.
In contrast, text-to-image models like Stable Diffusion~\citep{stable_diffusion} excel in visio-linguistic tasks, likely due to their text conditioning enhanceing semantic consistency in its cross-attention maps~\citep{li2023diffusion, clark2023texttoimage}. \citet{li2023diffusion} recently demonstrated this on the Winoground benchmark, reliably matching captions to images with fine-grained spatial differences using denoising diffusion scores (see~\Cref{diffusion_score}). Similar results have been shown for other text-to-image models, including Imagen~\citep{clark2023texttoimage}, with almost all of these methods outperforming CLIP variants on the same tasks.
%in Winoground's image-matching task.
%be used  lead to stronger performance on Winoground than CLIP variants. The denoising diffusion score is computed as the expectation of the gap between the predicted noise (conditioned on the text) and the noise added to the original image across multiple time-steps.
%\citet{clark2023texttoimage} has similarly shown that text-to-image generative models such as Imagen outperform CLIP on similar visio-linguistic reasoning tasks. 

While these works have shown the potential of using generative text-to-image models for visio-linguistic tasks, it remains computationally intensive. For instance, computing the denoising diffusion score for image-text matching involves multiple passes through a UNet model (approximately 892M parameters) with varying noise levels and time-steps. On an entry-level GPU, this can take up to a minute for a single image-text matching task, making it impractical for real-world and real-time applications. In contrast, CLIP models can classify images up to 18 times faster (see~\Cref{diffusion_score}), requiring only one pass through both image and text encoders. A promising research direction, therefore, lies in finding methods that combine the strong visio-linguistic capabilities of text-to-image models with the rapid inference of CLIP.
To this end, we introduce SDS-CLIP, a lightweight and sample-efficient fine-tuning approach for CLIP which distills knowledge from Stable Diffusion, and enhances CLIP's visio-reasoning capabilities.
%{\it Can we enhance CLIP's visio-linguistic capabilities by distilling knowledge from text-to-image generative models like Stable-Diffusion}? 
Specifically, we add a regularization term to CLIP's standard contrastive loss based on score-distillation sampling (SDS)~\cite{poole2022dreamfusion}.
This regularization encourages CLIP's embeddings to be aligned with the denoising diffusion loss from a text-to-image model.
%and differentiable image parameterizations~\cite{mordvintsev2018differentiable}.
By fine-tuning CLIP with this regularized objective on a small paired image-text dataset, specifically 118k image-text pairs from MS-COCO, we demonstrate an 1.5-7$\%$ performance gain compared to vanilla CLIP on Winoground and ARO, two highly challenging visio-linguistic reasoning benchmarks. Notably, this is achieved by only updating CLIP's LayerNorm parameters.
Furthermore, we show that SDS-CLIP's zero-shot performance is not impacted on a wide range of downstream datasets.

%CLIP's embeddings can be aligned with the denoising diffusion loss.
%With just around 118k image-text pairs from MS-COCO and fine-tuning only the LayerNorm parameters of CLIP, we demonstrate that SDS-CLIP achieves a 1.5-7$\%$ enhancement in visio-linguistic reasoning scores on benchmarks including Winoground and ARO. 
%Remarkably, this augmentation also slightly improves CLIP's zero-shot classification abilities.
%Our work demonstrates that internet-scale image-text contrastive models can be refined through a lightweight post-hoc fine-tuning step.
In summary, our contributions are as follows:
\begin{itemize}
    %\item We highlight the importance of the denoising diffusion loss from large-scale text-to-image models in visio-linguistic reasoning. 
    \item We introduce SDS-CLIP, a novel sample-efficient and parameter-efficient fine-tuning method that integrates a distillation-based regularization term from text-to-image models.
    \item We empirically validate our approach on challenging benchmarks and demonstrate an improvement in CLIP's visio-linguistic reasoning, without harming its zero-shot capabilities.
    %for {\it object-swap, relational} and {\it attribute-understanding} tasks. 
\end{itemize}
\vspace{-0.3cm}
\section{Denoising Diffusion Score for Visio-Linguistic Reasoning}
\label{sec: dds}
The Winoground benchmark establishes a challenging image-text matching task to measure a model's visio-linugistic reasoning abilities: given an image $x$, the model must match it with the correct caption $c^*$ from a set of captions $C = \{c_{i}\}_{i=1}^{n}$, where all caption contains the same words but each describes a different spatial arrangement of the objects, with only one being correct.
Concurrent works~\cite{clark2023texttoimage, li2023diffusion, krojer2023diffusion} to this paper have showed that it is possible to use the denoising diffusion score from text-to-image generative models to perform such an image-matching task. 
%These works find that this approach performs comparably to CLIP at zero-shot classification, but are much better than CLIP on relational and attribute-binding tasks which require compositional generalization. 
\begin{table*}[t!]
\centering
\scalebox{0.65}{
\begin{tabular}{SSSSSSS|SSS} \toprule
    {Model} & {\textbf{Wino-Overall}} & {Object} & {Relation} & {Both} & {1 Main Pred} & {2 Main Preds} & \textbf{ARO-Overall} & {ARO-Relation} & {ARO-Attribution}\\ \midrule
    \text{ViT-B/16(CLIP)}  & 0.24 & 0.28 & 0.18 & 0.57 & 0.29 & 0.11 & 0.57 & 0.52 & 0.62 \\ 
    {FT with $L_{CLIP}$} & 0.23 & 0.27 & 0.19 & 0.56 & 0.30 & 0.11 & 0.56 & 0.51 & 0.62 \\
   
    \rowcolor{LightCyan} \text{FT with $L_{CLIP} + L_{SDS}$}  & \textbf{0.31} & \textbf{0.35} & \textbf{0.25} & \textbf{0.69} & \textbf{0.36} & \textbf{0.16} & \textbf{0.58} & \textbf{0.535} & \textbf{0.63} \\ \midrule 
    \text{ViT-B/32(CLIP)}  & 0.30 & 0.35 & 0.22 & 0.80 & 0.34 & 0.18 & 0.55 & 0.50 & 0.61\\ 
    \text{FT with $L_{CLIP}$} & 0.28 & 0.31 & 0.20 & 0.76 & 0.31 & 0.16 & 0.55 & 0.50 & 0.60\\
   
    \rowcolor{LightCyan} \text{FT with $L_{CLIP} + L_{SDS}$}  & \textbf{0.32} & \textbf{0.38} & \textbf{0.23} & 0.69 & \textbf{0.36} & \textbf{0.20} & \textbf{0.575} & \textbf{0.53} & \textbf{0.62}\\ \midrule 
    \text{ViT-L/14(CLIP)}  & 0.28 & 0.27 & 0.25 & 0.57 & 0.29 & 0.24 & 0.57 & 0.53 & 0.61\\ 
    \text{FT with $L_{CLIP}$} & 0.26 & 0.27 & 0.25 & 0.56 & 0.30 & 0.23 & 0.57 &0.53 & 0.61\\

    \rowcolor{LightCyan}  \text{FT with $L_{CLIP} + L_{SDS}$}  & \textbf{0.295} & \textbf{0.32} & 0.25 & 0.53 & \textbf{0.32} & 0.18 & \textbf{0.595} &\textbf{0.55} & \textbf{0.64}\\ \midrule 
    \text{ViT-L/14-336(CLIP)}  & 0.27 & 0.32 & 0.21 & 0.57 & 0.30 & 0.19 & 0.57 & 0.53 & 0.61\\ 
    \text{FT with $L_{CLIP}$} & 0.23 & 0.28 & 0.19 & 0.53 & 0.26 & 0.17 & 0.57 & 0.53 & 0.61\\

   \rowcolor{LightCyan}  \text{FT with $L_{CLIP} + L_{SDS}$}  & \textbf{0.285} & \textbf{0.34} & \textbf{0.23} & 0.56 & \textbf{0.31} & \textbf{0.21} & \textbf{0.585} & \textbf{0.54} & \textbf{0.63}\\ \midrule 
    \text{ResNet-50(CLIP)}  & 0.25 & 0.29 & 0.19 & 0.5 & 0.27 & 0.18 & 0.58 & 0.53 & 0.63 \\ 
       \text{FT with $L_{CLIP}$} & 0.24 & 0.27 & 0.20 & 0.49 & 0.27 & 0.16 & 0.575 & 0.52 & 0.63\\

        \rowcolor{LightCyan} \text{FT with $L_{CLIP} + L_{SDS}$}  & \textbf{0.265} & \textbf{0.30} & \textbf{0.21} & 0.42 & \textbf{0.29} & \textbf{0.19} & \textbf{0.60} & \textbf{0.55} & \textbf{0.66}\\ \midrule 
\end{tabular}}
\caption{\textbf{Our fine-tuning method \vclip{} improves CLIP performance on the Winoground benchmark by 1.5$\%$ to 7$\%$ and upto 3$\%$ for the ARO-Relation and Attribution tasks across various CLIP variants}. Specifically, we find that our method improves on the sub-categories involving {\it object-swap} and {\it relational} understanding which comprise of the majority of the tasks in Winoground. Note that {\it only} fine-tuning with image-text pairs from MS-COCO without the distillation loss does not lead to any improvements. OpenCLIP results in~\Cref{open_clip_results}. \label{winoground_results}}
\vspace{-0.6cm}
\end{table*}
This can be formalized as follows: for an image $x$ and caption $c$, the denoising diffusion score,  denoted by $d(x, c)$, is defined as: 
\begin{equation}
    \label{ddscore_og}
    d(x,c) = \mathbb{E}_{t \sim T, \epsilon \sim \mathcal{N}(0,I)} [\| \epsilon_{\theta}(v_{\alpha}(x), t, c) - \epsilon \|^{2} ]
\end{equation}
This denoising diffusion score can then be used to select a correct caption $c^{*}$ from $C$ as:
\begin{equation}
    \label{ddscore}
    c^{*} = \arg \min_{c \in C} \mathbb{E}_{t \sim T, \epsilon \sim \mathcal{N}(0,I)} [\| \epsilon_{\theta}(v_{\alpha}(x), t, c) - \epsilon \|^{2} ]
\end{equation}
where $t$ is the sampled time-step, $\epsilon_{\theta}$ is the noise prediction UNet, $v_{\alpha}$ is an encoder (e.g., VQ-VAE) which maps the image $x$ to a latent code and $\epsilon$ is the sampled Gaussian noise.
Previous works~\citep{krojer2023diffusion} have demonstrated that by adopting this approach, text-to-image models performing strongly on visio-linguistic reasoning benchmarks like Winoground, outperforming contrastive models like CLIP by a significant margin (see~\Cref{diffusion_score}). For ARO, we obtain an accuracy of 0.63 with the diffusion score which is better than CLIP models.

\section{\vclip{}: Our Method}
The core idea of our approach is to regularize the contrastive objective in CLIP with the denoising diffusion score from Stable Diffusion (see ~Eq.(\ref{ddscore_og})). Our method builds on the recent work of~\cite{poole2022dreamfusion} which maps the output of a 3D NeRF model into the input space of Stable Diffusion's UNet and optimizes its parameteres with the denoising diffusion loss, also known as the score-distillation sampling (SDS). In a similar vein, we fine-tune the parameters of CLIP using SDS. Intuitively, our set-up can be viewed as a form of knowledge distillation where the teacher is the text-to-image model and the student is CLIP. As a result, in inference, CLIP can benefit from the visio-linguistic reasoning capabilities that are already learned by text-to-image diffusion models.
%Our approach augments CLIP's standard contrastive loss with a regularization term that is  on regularizing CLIP's contrastive objective using Stable Diffusion's denoising diffusion score (see Eq.(\ref{ddscore_og})). Inspired by works like Dreamfusion~\cite{poole2022dreamfusion}, which optimizes a 3D NeRF model with score-distillation sampling (SDS), we similarly fine-tune CLIP's parameters using SDS. This setup can be viewed as a form of knowledge distillation, where the teacher is a text-to-image model, and the student is CLIP. Consequently, in inference, CLIP can benefit from the visio-linguistic reasoning capabilities that are already learned by text-to-image diffusion models.

Formally, we map the output of CLIP's image encoder to the input space of Stable Diffusion's UNet. Specifically, we pass a given image $x$ through CLIP's image encoder $f_{\phi}$ and map its \texttt{<CLS>} embedding through a linear map $h_{w} \in \mathcal{R}^{d \times 4 \times 64 \times 64}$ into the input space of Stable Diffusion's UNet $\epsilon_{\theta}$. This can be formalized as $\epsilon_{\theta}(h_{w}(f_{\phi}(x)), t, c)$ where $t$ is the time step and $c$ is the corresponding text caption for the given image. We then use this term in place of $\epsilon_{\theta}(v_{\alpha}(x), t, c)$ in Eq. (\ref{ddscore}) to arrive as a denoising diffusion loss $L_{SDS}$ which encourages image-text binding with feedback from the diffusion loss: 
\begin{equation}
    \label{sds_loss}
    L_{SDS} =  \mathbb{E}_{t \sim T, \epsilon \sim \mathcal{N}(0,I)} [\| \epsilon_{\theta}(h_{w}(f_{\phi}(x)), t, c) - \epsilon \|^{2}
\end{equation}
We practically implement this by adding this $L_{SDS}$ loss to the original contrastive objective of CLIP such that it acts as a regularizer:
\begin{equation}
    L_{total} = L_{CLIP} + \lambda L_{SDS} 
\end{equation}
where $L_{CLIP}$ is defined in~\Cref{clip_objective} and $\lambda$ is a hyper-parameter that can be set with a grid search. We note that there are multiple ways to incorporate a diffusion loss into CLIP's objective. We found that as an additional loss term led to the best results, however, we include the full set of design choices we considered in the Appendix. 

%$f_{\phi}$ is the \texttt{<CLS>} embedding from the image-encoder of any CLIP like model which has been contrastively trained with image-text pairs.
Similar to differentiable image parameterizations~\cite{mordvintsev2018differentiable} where a given function is optimized by backpropogation through the image generation process, the UNet parameters $\theta$ are kept frozen during the optimization process. Specifically, given $L_{total}(\phi, \gamma, w, \theta)$:
\begin{equation}
    \phi*, \gamma*, w* = \min_{\phi, \gamma, w} L_{total}(\phi, \gamma, w, \theta)
\end{equation}
where $\phi$, $\gamma$, $w$ are the learnable parameters of CLIP's image encoder, text encoder and the linear map between CLIP and Stable Diffusion's UNet.
\begin{figure*}
    \hskip 0.2cm
  \includegraphics[width=15.8cm, height=5.1cm]{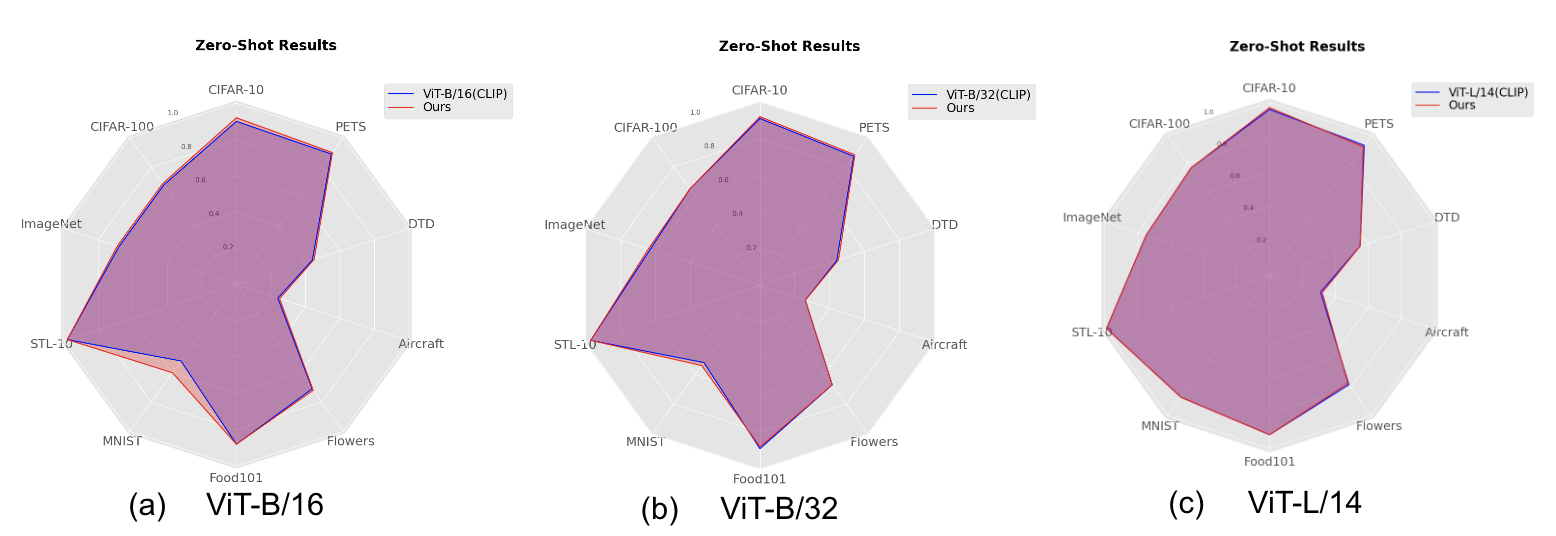}
    \caption{\label{clip_zeroshot} \textbf{Our fine-tuning method does not harm the zero-shot abilities of CLIP.} In fact for certain downstream datasets (e.g., ImageNet, CIFAR-10, MNIST, Aircraft) -- we observe an improvement in the zero-shot performance between $1\%-8\%$ for ViT-B/16. For other CLIP models, we find no drop in zero-shot performance. }%
    \vspace{-0.6cm}
\end{figure*}
%%%%%%%%%%%%%%%%%%%%%%%%%%%%%%%%%%%%%%%%%%%%%%%%%%%%%%%%%%%%%
\section{Experiments}
In this section, we empirically validate our proposed method \vclip{} on two types of tasks: i) visio-linguistic reasoning using two challenging benchmarks (Winoground, ARO) and ii) zero-shot image classification using a suite of downstream datasets (ImageNet, CIFAR-100, and others). Overall, we show that our method improves CLIP's performance significantly on Winoground and some key tasks in ARO, while also marginally improving downstream zero-shot classification performance.
\vspace{-0.1cm}
\subsection{Experimental Setup}
\textbf{CLIP Models. }  We consider the following CLIP variants in our experiments: (i) CLIP ViT-B/16; (ii) CLIP ViT-B/32; (iii) CLIP-ViT-L-14; (iv) CLIP-ViT-L-14 336px; (v) CLIP-ResNet-50. 
\textbf{Implementation Details. } Due to computational limit, we fine-tune CLIP from a publicly available checkpoint instead of training from scratch. Notably, we only fine-tune CLIP's LayerNorm parameters following~\cite{basu2023strong} along with the linear transformation $h_w$ -- accounting for only $\approx8M$ trainable parameters. We fine-tune these parameters using image-text pairs from MSCOCO~\cite{mscoco_cit}. In particular, we choose MSCOCO as it is relatively small and less noisy than other image-text datasets such as CC-12M~\cite{sharma-etal-2018-conceptual}. Both these factors make our fine-tuning method extremely sample- and parameter-efficient. 

\textbf{Baselines.} We compare our method with two different baselines: (i) pre-trained (vanilla) CLIP checkpoints; and (ii) CLIP fine-tuned on MS-COCO with the standard contrastive loss without the regularization term. 
\vspace{-0.4cm}
\subsection{Results}
\textbf{Winoground.} We evaluate \vclip{} on the challenging visio-linguistic reasoning benchmark, Winoground~\cite{thrush2022winoground}. In Table (\ref{winoground_results}), we find that our approach consistently improves performance across all Winoground sub-categories and CLIP variants, yielding absolute improvements ranging from 1.5$\%$ to 7$\%$. The largest gain of 7$\%$ is observed in ViT-B/16 (CLIP), with other CLIP variants showing consistent improvements of 1.5$\%$ to 2$\%$. In the Appendix(~\Cref{winoground_results_new}), we provide results for CLIP variants pre-trained on public data, where similar improvements are observed.
% We further analyze \vclip{}'s performance on Winoground sub-categories, including "object-swap," "relation," and "both," as well as based on the number of predicates in the captions. Our findings reveal consistent improvements in the "object-swap" and "relational" understanding sub-categories. On tasks with a single predicate in the captions, \vclip{} consistently enhances performance across all CLIP variants. However, on tasks with two predicates, \vclip{} improves performance for all CLIP variants except ViT-L/14. It's worth noting that tasks featuring both "object-swap" and "relation" tags constitute only approximately 5$\%$ of all Winoground tasks, which may not fully represent scenarios involving both object swaps and relational understanding.
On further inspection of the Winoground sub-categories, we find that \vclip{} shows consistent improvements in ``object-swap" and ``relation". 
It is worth noting that the ``both'' sub-category, which combines both ``object-swap" and ``relation" tags, makes up only $\tilde 5\%$ of all tasks, thus are potentially not fully representative of all scenarios involving both object swaps and relational understanding.
We also analyse \vclip's robustness to the number of predicates in captions and find that overall, it enhances performance in tasks where there are both one and two predicates.
\textbf{ARO.} The ARO dataset~\cite{yuksekgonul2023when} comprises tasks for (i) attribute-understanding and (ii) relational-understanding. In~\Cref{winoground_results}, we find that \vclip{} enhances performance by 1$\%$-3$\%$  in the "attribute-binding" and "relational understanding" tasks. %across all CLIP models. 

\textbf{Impact on CLIP's zero-shot performance.}
From~\Cref{clip_zeroshot}, we find that \vclip{}'s zero-shot classification capbilities are not impacted, relative to vanilla CLIP. In fact, we find that ViT-B/16's zero-shot performance improves across a range of downstream datasets (with up to 8$\%$ improvement for MNIST). 

While Stable-Diffusion is pre-trained on a much larger set of image-text pairs than CLIP, in~\Cref{scale_pretrain_data}, we show that the CLIP variants pre-trained on LAION-2B still suffer on Winoground. In fact, we show that using~\vclip{} can improve compositional reasoning of such CLIP variants. In~\Cref{ft_ccaptions}, we show results with fine-tuning on the larger CC-3M~\citep{sharma-etal-2018-conceptual}. 

% Taken together, these results highlight the effectiveness of our distillation strategy in improving CLIP's visio-linguistic reasoning, without any drop in zero-shot classification performance.
\vspace{-0.2cm}
%, across various CLIP models.
\section{Related Works}
While CLIP models~\citep{radford2021learning} are renowned for their robust zero-shot classification, recent research~\cite{thrush2022winoground, diwan2022winoground} has exposed their limitations in visio-linguistic reasoning. In contrast, recent studies have demonstrated that text-to-image  models~\cite{clark2023texttoimage, li2023diffusion, krojer2023diffusion, Chen2023RobustCV} outperform CLIP in  reasoning tasks. These models in fact leverage scores computed from the diffusion objective. We note that while~\citep{poole2022dreamfusion} use score-distillation sampling for text to 3D generation, ours is the first work to adapt the formulation as a regularizer and improve compositional abilities in CLIP.
%which have also proven effective in zero-shot classification. 
\vspace{-0.6cm}
\section{Conclusion}
\vspace{-0.2cm}
Our paper introduces \vclip{}, a novel data and parameter-efficient method that effectively enhances CLIP's visio-linguistic reasoning abilities by distilling knowledge from text-to-image models, without compromising its zero-shot abilities. 
%Notably, \vclip{} is highly parameter and data-efficient therefore enabling light-weight fine-tuning to improve compositionality. 
%This enhancement in visio-linguistic reasoning doesn't compromise the model's zero-shot performance, highlighting the benefits of knowledge distillation from text-to-image models to contrastive models.
%In summary, our work highlights the potential benefits of knowledge distillation from text-to-image models for improving reasoning in contrastive vision-language models. %especially in visio-linguistic reasoning tasks.
% Our paper demonstrates that distilling knowledge from text-to-image generative models into contrastive vision-language models such as CLIP significantly enhances CLIP's visio-linguistic reasoning abilities. Our method, \vclip{}, is remarkably sample- and parameter-efficient, fine-tuning just the LayerNorm parameters in CLIP using only 118k image-text pairs from MS-COCO. Importantly, this improvement in visio-linguistic reasoning does not compromise the model's downstream zero-shot performance. In summary, our work underscores the potential advantages of knowledge distillation from text-to-image models for enhancing contrastive vision-language models, particularly in visio-linguistic reasoning tasks.
\section{Limitations}
The primary limitation of our method is the inability to use large batch-sizes on moderate size GPUs. This is due to the fact that the regularizer $L_{SDS}$ requires a full backward pass through the UNet, even though its parameters are frozen. We also find that while the original diffusion score is good at {\it object-understanding, attribute-understanding} and {\it relational-understanding} tasks, it does not perform well on ordering tasks from the ARO dataset. For this reason, distillation from Stable-Diffusion potentially may not be effective in improving CLIP's performance on ordering tasks. Similar results are also observed in concurrent works such as~\cite{krojer2023diffusion}. 
\section{Ethical Considerations}
Vision-language models such as CLIP have been known for inheriting biases~\citep{DBLP:journals/corr/abs-2108-02818} due to their training data. Our work uses a well-known widely used dataset (MS-COCO) for the fine-tuning procedure and therefore does not introduce any additional bias. In fact, our distillation method mitigates some of the inherited bias in CLIP which earlier did not lead to good reasoning capabilities. 
% In our paper, we show that knowledge distillation from text-to-image generative models (e.g., Stable-Diffusion) to contrastive vision-language models such as CLIP can improve CLIP's visio-linguistic reasoning abilities on {\it object-swap, relational-understanding and attribute-binding} tasks. Our method for distillation -- \vclip{} is extremely light-weight and parameter-efficient, requiring only $\sim$118k training image-text pairs from MS-COCO and fine-tuning only the LayerNorm parameters in CLIP. Our empirical results also show that this improvement does not come at the cost of downstream zero-shot performance. In summary, our work provides evidence that distilling knowledge from strong text-to-image models can indeed be helpful in improving contrastive vision-language models, especially for visio-linguistic reasoning.
\bibliography{anthology,custom}

\begin{thebibliography}{27}
\expandafter\ifx\csname natexlab\endcsname\relax\def\natexlab#1{#1}\fi

\bibitem[{Agarwal et~al.(2021)Agarwal, Krueger, Clark, Radford, Kim, and Brundage}]{DBLP:journals/corr/abs-2108-02818}
Sandhini Agarwal, Gretchen Krueger, Jack Clark, Alec Radford, Jong~Wook Kim, and Miles Brundage. 2021.
\newblock \href {http://arxiv.org/abs/2108.02818} {Evaluating {CLIP:} towards characterization of broader capabilities and downstream implications}.
\newblock \emph{CoRR}, abs/2108.02818.

\bibitem[{Basu et~al.(2023)Basu, Massiceti, Hu, and Feizi}]{basu2023strong}
Samyadeep Basu, Daniela Massiceti, Shell~Xu Hu, and Soheil Feizi. 2023.
\newblock \href {http://arxiv.org/abs/2304.01917} {Strong baselines for parameter efficient few-shot fine-tuning}.

\bibitem[{Chen et~al.(2023)Chen, Dong, Wang, Yang, Duan, Su, and Zhu}]{Chen2023RobustCV}
Huanran Chen, Yinpeng Dong, Zhengyi Wang, X.~Yang, Chen-Dong Duan, Hang Su, and Jun Zhu. 2023.
\newblock Robust classification via a single diffusion model.
\newblock \emph{ArXiv}, abs/2305.15241.

\bibitem[{Clark and Jaini(2023)}]{clark2023texttoimage}
Kevin Clark and Priyank Jaini. 2023.
\newblock \href {http://arxiv.org/abs/2303.15233} {Text-to-image diffusion models are zero-shot classifiers}.

\bibitem[{Diwan et~al.(2022)Diwan, Berry, Choi, Harwath, and Mahowald}]{diwan2022winoground}
Anuj Diwan, Layne Berry, Eunsol Choi, David Harwath, and Kyle Mahowald. 2022.
\newblock \href {http://arxiv.org/abs/2211.00768} {Why is winoground hard? investigating failures in visuolinguistic compositionality}.

\bibitem[{Huang et~al.(2023)Huang, Tang, Chen, Zhang, Zhang, Chen, Zhao, Lv, Hu, and Zhang}]{huang2023structureclip}
Yufeng Huang, Jiji Tang, Zhuo Chen, Rongsheng Zhang, Xinfeng Zhang, Weijie Chen, Zeng Zhao, Tangjie Lv, Zhipeng Hu, and Wen Zhang. 2023.
\newblock \href {http://arxiv.org/abs/2305.06152} {Structure-clip: Enhance multi-modal language representations with structure knowledge}.

\bibitem[{Johnson et~al.(2016)Johnson, Hariharan, van~der Maaten, Fei{-}Fei, Zitnick, and Girshick}]{clevr}
Justin Johnson, Bharath Hariharan, Laurens van~der Maaten, Li~Fei{-}Fei, C.~Lawrence Zitnick, and Ross~B. Girshick. 2016.
\newblock \href {http://arxiv.org/abs/1612.06890} {{CLEVR:} {A} diagnostic dataset for compositional language and elementary visual reasoning}.
\newblock \emph{CoRR}, abs/1612.06890.

\bibitem[{Krojer et~al.(2023)Krojer, Poole-Dayan, Voleti, Pal, and Reddy}]{krojer2023diffusion}
Benno Krojer, Elinor Poole-Dayan, Vikram Voleti, Christopher Pal, and Siva Reddy. 2023.
\newblock \href {http://arxiv.org/abs/2305.16397} {Are diffusion models vision-and-language reasoners?}

\bibitem[{Li et~al.(2023)Li, Prabhudesai, Duggal, Brown, and Pathak}]{li2023diffusion}
Alexander~C. Li, Mihir Prabhudesai, Shivam Duggal, Ellis Brown, and Deepak Pathak. 2023.
\newblock \href {http://arxiv.org/abs/2303.16203} {Your diffusion model is secretly a zero-shot classifier}.

\bibitem[{Li et~al.(2022)Li, Li, Xiong, and Hoi}]{blip}
Junnan Li, Dongxu Li, Caiming Xiong, and Steven C.~H. Hoi. 2022.
\newblock \href {http://arxiv.org/abs/2201.12086} {{BLIP:} bootstrapping language-image pre-training for unified vision-language understanding and generation}.
\newblock \emph{CoRR}, abs/2201.12086.

\bibitem[{Lin et~al.(2014)Lin, Maire, Belongie, Bourdev, Girshick, Hays, Perona, Ramanan, Doll{\'{a}}r, and Zitnick}]{mscoco_cit}
Tsung{-}Yi Lin, Michael Maire, Serge~J. Belongie, Lubomir~D. Bourdev, Ross~B. Girshick, James Hays, Pietro Perona, Deva Ramanan, Piotr Doll{\'{a}}r, and C.~Lawrence Zitnick. 2014.
\newblock \href {http://arxiv.org/abs/1405.0312} {Microsoft {COCO:} common objects in context}.
\newblock \emph{CoRR}, abs/1405.0312.

\bibitem[{L{\"{u}}ddecke and Ecker(2021)}]{clip_seg_2}
Timo L{\"{u}}ddecke and Alexander~S. Ecker. 2021.
\newblock \href {http://arxiv.org/abs/2112.10003} {Prompt-based multi-modal image segmentation}.
\newblock \emph{CoRR}, abs/2112.10003.

\bibitem[{Minderer et~al.(2022)Minderer, Gritsenko, Stone, Neumann, Weissenborn, Dosovitskiy, Mahendran, Arnab, Dehghani, Shen, Wang, Zhai, Kipf, and Houlsby}]{minderer2022simple}
Matthias Minderer, Alexey Gritsenko, Austin Stone, Maxim Neumann, Dirk Weissenborn, Alexey Dosovitskiy, Aravindh Mahendran, Anurag Arnab, Mostafa Dehghani, Zhuoran Shen, Xiao Wang, Xiaohua Zhai, Thomas Kipf, and Neil Houlsby. 2022.
\newblock \href {http://arxiv.org/abs/2205.06230} {Simple open-vocabulary object detection with vision transformers}.

\bibitem[{Mokady et~al.(2021)Mokady, Hertz, and Bermano}]{mokady2021clipcap}
Ron Mokady, Amir Hertz, and Amit~H. Bermano. 2021.
\newblock \href {http://arxiv.org/abs/2111.09734} {Clipcap: Clip prefix for image captioning}.

\bibitem[{Mordvintsev et~al.(2018)Mordvintsev, Pezzotti, Schubert, and Olah}]{mordvintsev2018differentiable}
Alexander Mordvintsev, Nicola Pezzotti, Ludwig Schubert, and Chris Olah. 2018.
\newblock \href {https://doi.org/10.23915/distill.00012} {Differentiable image parameterizations}.
\newblock \emph{Distill}.
\newblock Https://distill.pub/2018/differentiable-parameterizations.

\bibitem[{Mu et~al.(2021)Mu, Kirillov, Wagner, and Xie}]{slip}
Norman Mu, Alexander Kirillov, David~A. Wagner, and Saining Xie. 2021.
\newblock \href {http://arxiv.org/abs/2112.12750} {{SLIP:} self-supervision meets language-image pre-training}.
\newblock \emph{CoRR}, abs/2112.12750.

\bibitem[{Poole et~al.(2022)Poole, Jain, Barron, and Mildenhall}]{poole2022dreamfusion}
Ben Poole, Ajay Jain, Jonathan~T. Barron, and Ben Mildenhall. 2022.
\newblock \href {http://arxiv.org/abs/2209.14988} {Dreamfusion: Text-to-3d using 2d diffusion}.

\bibitem[{Radford et~al.(2021{\natexlab{a}})Radford, Kim, Hallacy, Ramesh, Goh, Agarwal, Sastry, Askell, Mishkin, Clark, Krueger, and Sutskever}]{radford2021learning}
Alec Radford, Jong~Wook Kim, Chris Hallacy, Aditya Ramesh, Gabriel Goh, Sandhini Agarwal, Girish Sastry, Amanda Askell, Pamela Mishkin, Jack Clark, Gretchen Krueger, and Ilya Sutskever. 2021{\natexlab{a}}.
\newblock \href {http://arxiv.org/abs/2103.00020} {Learning transferable visual models from natural language supervision}.

\bibitem[{Radford et~al.(2021{\natexlab{b}})Radford, Kim, Hallacy, Ramesh, Goh, Agarwal, Sastry, Askell, Mishkin, Clark, Krueger, and Sutskever}]{clip_obj}
Alec Radford, Jong~Wook Kim, Chris Hallacy, Aditya Ramesh, Gabriel Goh, Sandhini Agarwal, Girish Sastry, Amanda Askell, Pamela Mishkin, Jack Clark, Gretchen Krueger, and Ilya Sutskever. 2021{\natexlab{b}}.
\newblock \href {http://arxiv.org/abs/2103.00020} {Learning transferable visual models from natural language supervision}.
\newblock \emph{CoRR}, abs/2103.00020.

\bibitem[{Rombach et~al.(2021)Rombach, Blattmann, Lorenz, Esser, and Ommer}]{stable_diffusion}
Robin Rombach, Andreas Blattmann, Dominik Lorenz, Patrick Esser, and Bj{\"{o}}rn Ommer. 2021.
\newblock \href {http://arxiv.org/abs/2112.10752} {High-resolution image synthesis with latent diffusion models}.
\newblock \emph{CoRR}, abs/2112.10752.

\bibitem[{Sharma et~al.(2018)Sharma, Ding, Goodman, and Soricut}]{sharma-etal-2018-conceptual}
Piyush Sharma, Nan Ding, Sebastian Goodman, and Radu Soricut. 2018.
\newblock \href {https://doi.org/10.18653/v1/P18-1238} {Conceptual captions: A cleaned, hypernymed, image alt-text dataset for automatic image captioning}.
\newblock In \emph{Proceedings of the 56th Annual Meeting of the Association for Computational Linguistics (Volume 1: Long Papers)}, pages 2556--2565, Melbourne, Australia. Association for Computational Linguistics.

\bibitem[{Thrush et~al.(2022)Thrush, Jiang, Bartolo, Singh, Williams, Kiela, and Ross}]{thrush2022winoground}
Tristan Thrush, Ryan Jiang, Max Bartolo, Amanpreet Singh, Adina Williams, Douwe Kiela, and Candace Ross. 2022.
\newblock \href {http://arxiv.org/abs/2204.03162} {Winoground: Probing vision and language models for visio-linguistic compositionality}.

\bibitem[{Wang et~al.(2021)Wang, Lu, Li, Tao, Guo, Gong, and Liu}]{clip_seg_1}
Zhaoqing Wang, Yu~Lu, Qiang Li, Xunqiang Tao, Yandong Guo, Mingming Gong, and Tongliang Liu. 2021.
\newblock \href {http://arxiv.org/abs/2111.15174} {{CRIS:} clip-driven referring image segmentation}.
\newblock \emph{CoRR}, abs/2111.15174.

\bibitem[{Xu et~al.(2023)Xu, Liu, Vahdat, Byeon, Wang, and Mello}]{xu2023openvocabulary}
Jiarui Xu, Sifei Liu, Arash Vahdat, Wonmin Byeon, Xiaolong Wang, and Shalini~De Mello. 2023.
\newblock \href {http://arxiv.org/abs/2303.04803} {Open-vocabulary panoptic segmentation with text-to-image diffusion models}.

\bibitem[{Yu et~al.(2022)Yu, Wang, Vasudevan, Yeung, Seyedhosseini, and Wu}]{yu2022coca}
Jiahui Yu, Zirui Wang, Vijay Vasudevan, Legg Yeung, Mojtaba Seyedhosseini, and Yonghui Wu. 2022.
\newblock \href {http://arxiv.org/abs/2205.01917} {Coca: Contrastive captioners are image-text foundation models}.

\bibitem[{Yuksekgonul et~al.(2023)Yuksekgonul, Bianchi, Kalluri, Jurafsky, and Zou}]{yuksekgonul2023when}
Mert Yuksekgonul, Federico Bianchi, Pratyusha Kalluri, Dan Jurafsky, and James Zou. 2023.
\newblock \href {https://openreview.net/forum?id=KRLUvxh8uaX} {When and why vision-language models behave like bags-of-words, and what to do about it?}
\newblock In \emph{The Eleventh International Conference on Learning Representations}.

\bibitem[{Zhong et~al.(2021)Zhong, Yang, Zhang, Li, Codella, Li, Zhou, Dai, Yuan, Li, and Gao}]{region_clip}
Yiwu Zhong, Jianwei Yang, Pengchuan Zhang, Chunyuan Li, Noel Codella, Liunian~Harold Li, Luowei Zhou, Xiyang Dai, Lu~Yuan, Yin Li, and Jianfeng Gao. 2021.
\newblock \href {http://arxiv.org/abs/2112.09106} {Regionclip: Region-based language-image pretraining}.
\newblock \emph{CoRR}, abs/2112.09106.

\end{thebibliography}
\newpage 
\clearpage 
\appendix
\section{Benchmark Datasets}
\label{benchmark_datasets}
\subsection{Benchmark datasets}
Winoground~\cite{thrush2022winoground, diwan2022winoground} is a challenging vision-language dataset for evaluating the visio-linguistic characteristics of contrastively trained image-text models. The dataset consists of 400 tasks, where each task consists of two image-text pairs. The objective is to independently assign the correct text caption to each image. Each task is also annotated with meta-data corresponding to whether the task requires object-understanding, relational-understanding or both. The tasks in Winoground are challenging as the images differ in fine-grained ways and assigning the correct text captions requires inherent compositional visual reasoning.
%As a testament to its challenging nature, the human baseline on Winoground is only $\sim89\%$. 

%% Zou Lab dataset
ARO~\cite{yuksekgonul2023when} similarly tests visio-linguistic reasoning and consists of three types of tasks: (i) Visual Genome Attribution to test the understanding of object properties; (ii) Visual Genome Attribution to test for relational understanding between objects; and (iii) COCO-Order and Flickr30k-Order to test for order sensitivity of the words in a text, when performing image-text matching. We highlight that Winoground though slightly smaller in size than ARO is more challenging as it requires reasoning beyond visio-linguistic compositional knowledge~\cite{diwan2022winoground}. 

\subsection{Does distilling features directly from  UNet help?}
Previous works such as~\cite{xu2023openvocabulary} find that the frozen features of the UNet contain structural information about the image. Motivated by this, we also investigate if distilling knowledge directly from the frozen UNet features is beneficial, Given an image $x$ and its caption $c$, the frozen features $f$ from the UNet (where $I(x,c) = \epsilon_{\theta}(v_{\alpha}(x), t, c)$, similar to~\cite{xu2023openvocabulary}) can be extracted. We then use these frozen internal representations from the UNet to regularize features of the image encoder in CLIP. In particular:
\begin{equation}
    L_{total} = L_{CLIP} + \lambda \|h_{w}(f_{\phi}(x) - I(x,c)) \|_{2}^{2}
\end{equation}
However, we find that distillation in this way does not lead to improved performances for visio-linguistic reasoning. In fact, for ViT-B/16 (CLIP) we find the Winoground score to decrease from 0.24 to 0.23. This result shows that using score-distillation sampling which involves backpropogation through the UNet is critical to distill knowledge from diffusion models to other discriminative models.
\section{SDS-CLIP: Algorithm}
\begin{algorithm}[ht]
\caption{Algorithm to fine-tune CLIP with distillation from Stable-Diffusion for improved visio-linguistic reasoning}\label{alg:cap}
\begin{algorithmic}
\Require $\mathcal{D}$: \text{image-text pairs}, $f_{\phi}$: CLIP's image-encoder, $g_{\gamma}$: CLIP's text-encoder, $\epsilon_{\theta}$: UNet; N: Number of Epochs; $\lambda$: Hyper-parameter for the regularizer; $|B|$: Batch-size.
% \Ensure $y = x^n$
% \State $y \gets 1$
% \State $X \gets x$
% \State $N \gets n$
\While{$i \neq N$}
\State $\{x_{j}, y_{j}\}_{j=1}^{|B|} \gets$ \text{Sample a batch from $\mathcal{D}$}
\State $t \gets$ \text{Sample time-steps using DDPM}
\State $\epsilon \gets$ \text{Sample Gaussian noise $\epsilon \sim \mathcal{N}$(0, I) }
\State $L_{clip \gets}$ \text{Compute contrastive loss as in~\cref{contrastive_main}}
\State $L_{SDS} \gets$ \text{Compute SDS loss as in~\cref{sds_loss}}
\State $L_{total} \gets L_{clip} + \lambda L_{SDS}$
\State $L_{total}$\text{.backward()} \Comment{Backprop}
\State $\phi, \gamma, w \gets $ \text{Update the relevant parameters}
% \If{$N$ is even}
%     \State $X \gets X \times X$
%     \State $N \gets \frac{N}{2}$  \Comment{This is a comment}
% \ElsIf{$N$ is odd}
%     \State $y \gets y \times X$
%     \State $N \gets N - 1$
% \EndIf
\State $i \gets i + 1$
\EndWhile
\end{algorithmic}
\end{algorithm}
\section{Preliminaries}
\subsection{CLIP}
\label{clip_objective}
CLIP~\cite{clip_obj} is a image-text model which is pre-trained using a contrastive objective, typically on internet-scale data. The core intuition of the training objective is to align the text and image embeddings of image-text pairs in a shared embedding space. To do this, CLIP consists of two components: (i) an image encoder $f_{\phi}$ which transforms a raw image $x_{i}$ into an image embedding $e_{img}(x_{i}) = f_{\phi}(x_{i}) \in \mathbb{R}^{d}$, also denoted by the \texttt{<CLS>} token; and (ii) a text encoder $g_{\gamma}$ which transforms a raw text caption $c_{i}$ into a text embedding $e_{text}(c_{i}) = g_{\gamma}(c_{i}) \in \mathbb{R}^{d}$ also denoted by \texttt{<EOS>} token, both of which map to an embedding dimensionality d. Given a dataset $\mathcal{D} = \{(x_{i},c_{i})\}_{i=1}^{N}$ of image-text pairs, where $(x_{i}, y_{i})$ is the $i^{th}$ image-text pair, CLIP uses a contrastive objective to pull the image and text embeddings of matched pairs together, while pushing those of unmatched pairs apart.
%\{ e_{img}(x_{i}),e_{text}(c_{i})\}_{i=1}^{N}$ to pull the embeddings of paired image-text pairs while pushing apart the embeddings of unmatched pairs.
Formally, the contrastive objective can be defined as:
\begin{equation}
    \label{contrastive_main}
    L_{CLIP} = L_{image-text} + L_{text-image}
\end{equation}
where:
\tiny
\begin{equation}
    L_{image-text} = -\frac{1}{2N}\sum_{j=1}^{N} \log \{\frac{\exp(e_{img}(x_{j})^{T}e_{text}(c_{j})/\tau)}{\sum_{k=1}^{N}\exp((e_{img}(x_{j})^{T}e_{text}(c_{k})/\tau))} \}
\end{equation}
\normalsize
\tiny
\begin{equation}
    L_{text-image} =  -\frac{1}{2N}\sum_{j=1}^{N} \log \{\frac{\exp(e_{img}(x_{j})^{T}e_{text}(c_{j})/\tau)}{\sum_{k=1}^{N}\exp((e_{img}(x_{k})^{T}e_{text}(c_{j})/\tau))} \}
\end{equation}
\normalsize
where $\tau$ is a trainable temperature parameter. Usually $\mathcal{D}$ is an internet-scale dataset consisting of millions of image-text pairs. Furthermore, during pre-training, the embeddings $e_{img}(x_{i})$ and $e_{text}(c_{i})$ are normalized to have a unit-norm.

\section{When does distillation not help CLIP?} 
\label{failure_mode_sd}
While we find that distilling knowledge from Stable-Diffusion to CLIP helps in {\it object-swap}, {\it relational-understanding} and {\it attribution-binding} visio-linguistic tasks, it does not help on tasks where the order of the text is perturbed (e.g. the COCO-Order and Flickr-Order tasks in the ARO dataset). 
%This is shown in the final two columns of Table (\ref{aro_results}).
In fact, we find that the denoising diffusion score in~\Cref{ddscore_og} leads to accuracies of 0.24 for COCO-Order and 0.34 for Flickr-Order which is in fact lower than CLIP models. Concurrent works~\cite{krojer2023diffusion} has shown similarly low performance for text-ordering tasks. A potential reason could be that ordering tasks only test for grammatical understanding which current text encoders cannot effectively model. Another reason could be that the denoising diffusion score is not affected by word ordering as the image semantics are not changed as a result. 
% \begin{figure}
%     \hskip 0.1cm
%   \includegraphics[width=7.9cm, height=5.6cm]{bar_time.png}
%     \caption{\label{clip_time} \textbf{Denoising Diffusion Score computation takes $\sim$ 40x more time than the image-text alignment score in CLIP.} The higher inference time incurred by diffusion score computation from text-to-image generative models such as Stable-Diffusion make it infeasible to be usable in practice. }%
% \end{figure}
\section{Notes on Fine-tuning Dataset}
We use MS-COCO~\citep{mscoco_cit} which is widely used for multimodal learning. This dataset does not contain any names or uniquely identifies individual people or offensive content.
\section{More Experimental Details}
\textbf{Hyper-parameters.} We perform a hyperparameter sweep for the learning rate and the regularization hyperparameter $\lambda$ for ViT-B/16. We use these same hyperparameters for different CLIP variants including ViT-B/32, ViT-B/14, ViT-L/14-336px and ResNet-50. In particular, we set $\lambda = 0.001$ and set the learning rate as $5 \times 10^{-5}$. We use a batch-size of 32 for all the different CLIP models. We use Stable-Diffusion v1-4 as the teacher model in our experiments. 

\textbf{Note on Full Fine-tuning. } All our experiments were primarily done by fine-tuning only the LayerNorm parameters. In the initial phase of the project, we also fine-tune all the parameters of the text and image encoder in CLIP, however it results in worse performances than those reported in Table. (\ref{winoground_results}). Potentially, this can be due to overfitting issues when used in conjunction with the new regularizer. We therefore run all the experiments with LayerNorm tuning as it leads to the best results. 

\textbf{Total GPU Hours.} For all our experiments we use NVIDIA-A6000 and each fine-tuning experiment takes $\approx$6 hours. 
\begin{table}[t!]
\centering
\scalebox{0.48}{
\begin{tabular}{SSSSSSS} \toprule
    {Model} & {\textbf{Overall}} & {Object} & {Relation} & {Both} & {1 Main Pred} & {2 Main Preds}\\ \midrule
    \text{ViT-B/16(LAION 400M)}  & 0.24 & 0.29 & 0.17 & 0.59 & 0.28 & 0.11  \\ 
    \text{COCO FT with $L_{CLIP}$} & 0.24 & 0.26 & 0.21 & 0.54 & 0.31 & 0.10 \\
    \text{COCO FT with $L_{CLIP} + L_{SDS}$}  & \textbf{0.30} & \textbf{0.34} & \textbf{0.23} & 0.55 & \textbf{0.33} & \textbf{0.14} \\ \midrule 
\end{tabular}}
\caption{\textbf{Additional results on Winoground with ViT-B/16 CLIP pre-trained on public data (LAION-400M)}. \label{winoground_results_new}}
\vspace{-0.0cm}
\end{table}
\section{Additional Results with Stable-Diffusion-v2-1}
\label{sd_2_1}
In particular, with our distillation strategy with Stable-Diffusion v-2.1 as a teacher – we obtain the following results on Winoground: (i) ViT-B/16: 0.35; (ii) ViT-B/32: 0.33; (iii) ViT-L/14: 0.31; (iv) ViT-L/14-336px: 0.31; (iv) ResNet-50: 0.28; All the scores are higher than the fine-tuned model with Stable-Diffusion-v1-4 as the teacher, therefore highlighting that a teacher with better compositional generation capabilities will be a better choice. 
\section{Fine-tuning with Conceptual Captions}
\label{ft_ccaptions}
We primarily use MS-COCO as : (i) It's a relatively small dataset which can keep the fine-tuning steps relatively smaller and scaling the fine-tuning dataset will increase fine-tuning time; (ii) It's a well-established, relatively diverse and well annotated image-text dataset which is used by the community. We also fine-tuned with CC-3M~\citep{sharma-etal-2018-conceptual}, but found the improvements to be similar in lines to that using MS-COCO. For e.g., On Winoground with CC-3M, we find the following performance after distillation with Stable-Diffusion-v1-4: (i) ViT-B/16: 0.32; (ii) ViT-B/32: 0.32; (iii) ViT-L/14: 0.30; (iv) ViT-L/14-336px: 0.28; (iv) ResNet-50: 0.27. 
These scores are only marginally better than using MS-COCO, although the dataset size is more than 30 times -- which shows that a high-quality dataset such as MS-COCO is sufficient for improving compositional abilities in CLIP. 
\section{Results with OpenCLIP}
\label{open_clip_results}
In~\Cref{winoground_results_new}, we show that our method is compatible with OpenCLIP. In particular, we find that distillation to OpenCLIP improves its visio-linguistic score from 0.24 to 0.30. These results highlight the generalizability of our distillation method. 
\section{Additional Results on CLEVR}
\label{clevr_results}
We apply our fine-tuned model on the CLEVR task~\citep{clevr} – which consists of images of 3D shapes isolating phenomena such as spatial reasoning or attribute binding. We find that the diffusion-score leads to a score of 0.67, whereas the best CLIP variant in our test-bed (CLIP ViT-L/14) scored 0.63. With our distillation loss during fine-tuning – this score improved to 0.65 with a 2$\%$ gain. 
\section{Is it the Scale of Pre-Training Data Which Helps? }
\label{scale_pretrain_data}
\begin{table}[t!]
\centering
\scalebox{0.48}{
\begin{tabular}{SSSSSSS} \toprule
    {Model} & {\textbf{Overall}} & {Object} & {Relation} & {Both} & {1 Main Pred} & {2 Main Preds}\\ \midrule
    \text{ViT-B/16(LAION 2B)}  & 0.27 & 0.32 & 0.19 & 0.61 & 0.29 & 0.12  \\ 
    \text{COCO FT with $L_{CLIP} + L_{SDS}$}  & \textbf{0.31} & \textbf{0.36} & \textbf{0.24} & 0.53 & \textbf{0.36} & \textbf{0.17} \\ \midrule 
\end{tabular}}
\caption{\textbf{CLIP (Pre-trained with 2B images) still underperforms on Winoground.} We show the CLIP even when trained with LAION-2B (similar scale of training data as Stable-Diffusion) still underperforms the diffusion score from Stable-Diffusion. This shows that scale of data alone cannot be useful in mitigating reasoning capabilities in CLIP. \label{laion_2b}
}
\vspace{-0.0cm}
\end{table}
In~\Cref{laion_2b}, we show that CLIP models even when trained at the same scale of pre-training data as Stable-Diffusion (LAION-2B) struggle on the Winoground dataset. We specifically highlight that CLIP (when pre-trained on 2B image-text pairs) obtain a score of 0.27, whereas the diffusion model when trained on similar pre-training corpus obtains a score of 0.35. This clearly shows that at a similar pre-training scale, diffusion models (with their diffusion objective) are better compositional learners than CLIP like models. 
Our distillation method from Stable-Diffusion improves the Winoground score from 0.27 to 0.31 on CLIP(pre-trained on 2B image-text pairs).

\section{Beyond CLIP}
We find that Open-CoCa~\citep{yu2022coca} pre-trained on 2B image-text pairs obtains a score of 0.30 on Winoground. With our distillation strategy, we find that the score improves to 0.33 highlighting that our distillation strategy can be used for models beyond CLIP. A full investigation of the impact of our distillation method on various vision-language models is deferred towards future work.
\end{document}